\documentclass{article}

\usepackage{url}
\usepackage{booktabs}

\begin{document}

\title{A review of sentiment computation methods with R packages}

\author{Maurizio Naldi\\University of Rome Tor Vergata\\Dpt. of Civil Engineering and Computer Science\\Via del Politecnico 1, 00133 Rome, Italy}
\date{January 2019}
\maketitle




\begin{abstract}
Four packages in R are analyzed to carry out sentiment analysis. All packages allow to define custom dictionaries. Just one - SentimentR - properly accounts for the presence of negators.  
\end{abstract}

\section{Introduction}
Sentiment analysis is an increasing area of research and application, due to the sheer size of unstructured data that is now available \cite{feldman2007text,gandomi2015beyond}. Numerous textbooks have been published that illustrate the major algorithms to be used for text mining, and specifically for sentiment analysis \cite{liu2015sentiment,cambria2017practical,feldman2007text}.

The languages of choice for sentiment analysis appear to be Python and R. In particular some books have been devoted to the use of R for these applications  \cite{kumar2016mastering} \cite{kwartler2017text}. However, several packages are available that provide built-in functions for sentiment computation in the R framework. Those functions allow data analysts to get rid of the burden of writing basic code to perform sentiment computation, at least at a first level of analysis, but differ for their breadth and accuracy, so that a problem of choosing the most suitable package exists.

In this paper, we provide a survey of those packages, highlighting their characteristics and critical issues, so as to help the prospective user to get a panorama view of the available libraries and make an informed choice.

\section{The packages}

The packages we consider in this survey are those including specific functions to compute sentiment. This sentiment analysis can be applied at several levels: a single sentence; the set of sentences making up a review; the set of reviews for a specific product/service (in the following, we imagine the text of interest to be the set of customer reviews appearing on websites such as Amazon or Trip Advisor). For each item in the group under analysis, the packages provide a numeric score, whose sign tells us the sentiment polarity and whose absolute value tells us the intensity of the sentiment.

The packages under examination in this paper are:
\begin{itemize}
    \item syuzhet;
    \item Rsentiment;
    \item SentimentR;
    \item SentimentAnalysis.
\end{itemize}

These packages share a common feature: they adopt the bag-of-words approach, where the sentiment is determined on the basis of the individual words occurring in the text, neglecting the role of syntax and grammar. The words appearing in the text are compared against one or more lexicons where positive and negative words are listed and typically associated with a degree of intensity (of positiveness or negativeness). The number of matches between the words in the text and the lexicon(s) as well as the degrees of intensity of the sentiment associated to those words are considered to arrive an an overall numeric score representing the overall sentiment at the level of interest.

In the next sections, we describe those packages, the associated lexicons, and the functions that each of them employs to compute the sentiment of a sentence (or a set of sentences).

\section{The \textit{syuzhet} package}

This package is available on CRAN (the Comprehensive R Archive Network) at the URL \url{http://cran.r-project.org/web/packages/syuzhet/}. A list of the functions embedded in the package is available at \url{https://cran.r-project.org/web/packages/syuzhet/syuzhet.pdf}. It has been applied to several sources of data: reviews on TripAdvisor \cite{valdivia2017sentiment}, German novels to predict happy endings \cite{zehe2016prediction}, tweets to detect trending sentiments in political elections \cite{kolagani2017identifying}.

Syuzhet allows us to choose among four sentiment lexicons or define a custom lexicon of our own (since version 2.0\footnote{see the page maintained by Matthew Jockers \url{https://cran.r-project.org/web/packages/syuzhet/vignettes/syuzhet-vignette.html}}). The four standard lexicons are:
\begin{itemize}
    \item syuzhet;
    \item afinn;
    \item bing;
    \item nrc.
\end{itemize}

All those lexicons can be retrieved within the \textit{syuzhet} package through a function (see below for details), which returns a data frame with two columns, reporting respectively the words and their sentiment score.

The \textit{syuzhet} lexicon is the default one and was developed in the Nebraska Literary Lab under the direction of Matthew L. Jockers. It comprises 10748 words with an associated sentiment value, spanning the $[-1,1]$ range (actually 16 values in it). Negative words (i.e., with a negative sentiment value) dominate, since they account for 7161 out of 10748; positive words are of course the remaining 3587. The lexicon itself can be retrieved through the function \texttt{get\_sentiment\_dictionary()}. 

The \textit{afinn} lexicon was developed by Finn Arup Nielsen as the \textit{AFINN Word Database}. It includes Internet slang and obscene words. It has been built starting from a set of obscene words and gradually extended by examining Twitter postings and sets of words as extracted from  the Urban Dictionary and Wiktionary to also include acronyms and abbreviations  \cite{nielsen2011new}.  The resulting lexicon is made of 2477 words (then much smaller than the  \textit{syuzhet} one). Again, the number of negative words is much larger: 1598 negative words vs 878 positive ones, and a single neutral word ("some kind"). The score range is $[-5,5]$ (11 values in that range), hence much wider than that employed in the \textit{syuzhet} lexicon. 
The lexicon can be retrieved through the function \texttt{get\_sentiment\_dictionary("afinn")}.

The \textit{bing} lexicon was developed by Minqing Hu and Bing Liu as the \textit{Opinion Lexicon}. It is made of 6789 words, of which 2006 are positive and 4783 are negative. The whole lexicon can be retrieved as a zipped file on the webpage \url{https://www.cs.uic.edu/~liub/FBS/sentiment-analysis.html#lexicon}. Aternatively, it can be  retrieved through the function \texttt{get\_sentiment\_dictionary("bing")} within the \textit{syuzhet} package. Score assignment is rather sharp: it is either -1 or 1.

The information about the three lexicons is summarized in Table \ref{tab:syuzhet} for comparison. If we consider size and resolution (i.e., the capability of resolving different grades of sentiment by using multiple values), we can note that the \textit{afinn} lexicon is the smallest of the three, while the \textit{bing} one exhibits the minimum resolution. The best lexicon under both criteria (size and resolution) is \textit{syuzhet}.

\begin{table}[]
    \centering
    \begin{tabular}{lcccc}
    \toprule
    Lexicon     & No. of words & No. of positive words & No. of negative words & Resolution\\
    \midrule
    Syuzhet & 10748 & 3587 & 7161 & 16\\
    Afinn & 2477 & 878 & 1598 & 11\\
    Bing & 6789 & 2006 & 4783 & 2\\
    \bottomrule
    \end{tabular}
    \caption{Lexicons in the \textit{syuzhet} package}
    \label{tab:syuzhet}
\end{table}

The \textit{nrc} lexicon is a bit different from those considered so far. Instead of just being concerned with the polarity, i.e. reporting positive or negative words, it assigns a sentiment type, using the following 8 additional categories:
\begin{itemize}
    \item anger;
    \item anticipation;
    \item disgust;
        \item fear;
        \item joy;
        \item sadness;
        \item surprise;
        \item trust.     
\end{itemize}
As to its size, this lexicon comprises 13889 words, distributed among the different categories as shown in Table \ref{tab:nrc}.

\begin{table}[]
    \centering
    \begin{tabular}{lc}
    \toprule
      Category   &  No. of words\\
      \midrule
      Anger & 1247\\
    Anticipation     & 839\\
    Disgust & 1058\\
    Fear & 1476\\
    Joy & 689\\
    Sadness  & 1191\\
    Surprise & 534\\
    Trust & 1231\\
    Positive & 2312\\
    Negative & 3324\\
    \bottomrule
    \end{tabular}
    \caption{Words in the \textit{nrc} lexicon}
    \label{tab:nrc}
\end{table}

The sentiment can be computed for each sentence through the \texttt{get\_sentiment} function. It applies to a character vector, and the lexicon to be used is to be specified. The sentiment computation is rather straightforward. It just finds all the lexicon words contained in each element of the input vector and computes the algebraic sum of the pertaining sentiment values. For example, for the sentence "This device is perfect but noisy", using the \textit{afinn lexicon}, the \texttt{get\_sentiment} function finds the two words "perfect" and "noisy" in that lexicon, with sentiment values +3 and -1 respectively, so that it outputs the overall sentiment score $3-1=2$. The same sentence analysed through the \textit{bing} lexicon would get a 0 score, since the two words are again the only ones found in that lexicon, but with sentiment values +1 and -1 respectively.

As to the \textit{nrc} lexicon, the occurrence in the text of a word appearing in one of the categories (say the disgust one) counts as 1 in the sentiment score for that category. So, if a sentence contains 3 words listed in the list of words for disgust, the score for that sentence in the disgust category will be 3. When using the \textit{nrc} lexicon, rather than receiving  the algebraic score due to positive and negative words, each sentence gets a score for each sentiment category. 

A major problem of this package is that it does not properly consider negatives. In the \textit{bing} lexicon, the word "not" is not included among the negative words, and the score obtained for the sentence "This device is not perfect" is exactly the same as "The device is perfect", i.e., 1 as if the negative were not there. The same can be said of the \textit{syuzhet} lexicon, which does not contain negatives in its word list. We have a marginally better situation in the \textit{afinn} lexicon, where the negative is recognized and dealt with just in the expressions "does not work", "not good", and "not working", which are included in the lexicon as negative words.



\section{The \textit{Rsentiment} package}
The package is available at \url{https://cran.r-project.org/web/packages/RSentiment/index.html}, while a description of its functions can be retrieved at \url{https://cran.r-project.org/web/packages/RSentiment/RSentiment.pdf}. It was written by Subhasree Bose with contributions from Saptarsi Goswami, though the paper describing it reports multiple authors \cite{Subhasree}. It has been employed to perform sentiment analysis in several contexts: driving systems for cars \cite{biondi2017partial}, political preferences through the analysis of tweets \cite{kassraie2017election}, transportation systems again through the analysis of tweets \cite{haghighi2018using}.

The algorithm uses Parts of Speech (PoS) tagging to tag each word in the sentence. In particular, it is reported to use a set of cases of sequence of occurrence of various parts of speech. For example, it checks if there is any adverb or adjective bearing positive score following any negative quantifier, such as "not", "no", "none", and "never", and accordingly assigns a score to it.

The package provides three functions that can be used to score sentences:
\begin{itemize}
    \item \texttt{calculate\_score}, which outputs a text description of the score;
    \item \texttt{calculate\_sentiment}, which classifies a sentence or a text into a sentiment category;
    \item \texttt{calculate\_total\_presence\_sentiment}, which returns the total number of sentences in each sentiment category.
\end{itemize}

The score is provided as either a positive or negative number, reporting the algebraic sum of the number of positive and negative words found in the text (hence each positive word is given a score of 1, and each negative word is given a score of -1). For example, the scores assigned to the sentences "This device is good", "This device is very good", "This device is very very good", and "This device is good but bad" (sorry for the contradiction) are respectively 1,2, 3, and 0. In many cases the introduction of negations (such as "not") makes the program crash. In the case of sarcasm the score is 99. However, the default lexicon does not appear to be publicly available.

The package classifies sentences into the following 6 categories:
\begin{itemize}
    \item Very Negative;
    \item Negative;
    \item Neutral;
    \item Positive;
    \item Very Positive;
    \item Sarcasm.
\end{itemize}
The thresholds to classify sentences into those categories are very simple: sentences are very negative if their score is smaller than -1, and and very positive if their score is larger than 1; sentences are instead simply negative if their score is -1 and positive if their score is exactly 1; neutral sentences have of course a zero score.

Though the package employs its own lexicon, a very good feature of \textit{Rsentiment} is that it allows to use one's own lexicons. These must be provided as separate lists of positive and negative words. The functions to be used in this case are \texttt{calculate\_custom\_score}, \texttt{calculate\_custom\_sentiment}, and \texttt{calculate\_custom\_total\_presence\_sentiment} respectively to get the score, the sentiment as a text, and the number of sentences per category.

The package is however reported to exhibit several issues. A critical issue is the installation, wince it requires Java and that may require special attention on Apple devices (see, e.g., \url{https://stackoverflow.com/questions/48580061/java-issue-with-rsentiment-package}. Also, some problems in recognizing the correct meaning of sentences including negatives have been reported, e.g., on the very blog of the author at \url{https://fordoxblog.wordpress.com/2017/03/12/rsentiment/}.

\section{The \textit{SentimentR} package}
The SentimentR package was developed by Tyler Rinker. It is available at \url{https://cran.r-project.org/web/packages/sentimentr/sentimentr.pdf}. The package has been employed in a number of contexts: to predict stock prices from Google news \cite{chen2015predicting}; to analyze sentiments expressed on Twitter by energy consumers \cite{ikoro2018analyzing}; and to analyze the sentiment of patients with critical illness \cite{weissman2019construct}.

It adopts a dictionary lookup approach that tries to incorporate weighting for valence shifters (negators and amplifiers/deamplifiers, which respectively reverse, increase, and decrease the impact of a polarized word). Though its aim is to improve the polarity recognition performance with respect to the \textit{syuzhet} package, which does not recognize valence shifters, it does so at the expense of speed. Though the author has striven to balance accuracy and speed, the SentimentR polarity computation may be slower on large datasets.

The importance of valence shifters and other modifiers can be assessed by taking a look at Table \ref{tab:modifiers}, where the occurrence of valence shifters is reported for a sample of texts.  Negators appear to occur around 20\% of the time a polarized word appears in a sentence in most texts. Amplifiers appear with a slightly lower frequency, while deamplifiers are much rarer. instead, adversative conjunctions appear with polarized words around 10\% of the time in most cases.

\begin{table}[]
    \centering
    \begin{tabular}{lcccc}
    \toprule
   Text	& Negator & Amplifier	& Deamplifier & Adversative\\
   \midrule
Canon reviews & 21\% & 23\%	& 8\%	 & 12\%\\
2012 presidential debate & 23\%	& 18\% & 1\% & 11\%\\
Trump speeches	& 12\% & 14\% & 3\% & 10\%\\
Trump tweets & 19\%	& 18\%	& 4\%	& 4\%\\
Dylan songs	& 4\%	& 10\%	& 0\%	& 4\%\\
Austen books & 21\%	& 18\%	& 6\%	&11\%\\
Hamlet & 26\% &	17\%	& 2\%	&16\%\\    
\bottomrule
\end{tabular}
    \caption{Occurrence of modifiers}
    \label{tab:modifiers}
\end{table}

The lexicon can be retrieved through the \texttt{show(lexicon::hash\_sentiment\_jockers\_rinker)} command. It contains 11709 words, whose individual scores may take one of 19 values in the $[-2,1]$ range. The package employs that lexicon in combination with a group of 140 valence shifters, which can be retrieved through the \texttt{show(lexicon::hash\_valence\_shifters)} command. The valence shifters take an integer value from 1 to 4. Emoticons can be replaced by their word equivalent through the \texttt{replace\_emoticon($\cdot$)} command, so as to be included in the score computation. However, its own dictionaries can be modified, and new custom dictionaries (\textit{keys} in the package lingo) can be built. Words can be added to or removed from an existing dictionary through the \textit{update\_key} function, using respectively the \textit{data.frame} or the \textit{drop} option. In order to create a new dictionary, a two-column data frame has to be created, with words in the first column and its value on the second one. The new dictionary has to be declared within the \textit{sentiment()} function through the \textit{polarity\_dt} option.

This package allows to compute the polarity both of a single sentence and a full text. The functions for those purposes are respectively \texttt{sentiment()} and \texttt{sentiment\_by()}. 

Both functions return a dataframe with four columns:
\begin{enumerate}
    \item \texttt{element\_id}, which is the ID/Serial Number of the given text;
    \item \texttt{sentence\_id}, which is the ID/Serial Number of the sentence and is equal to \texttt{element\_id} in the case of \texttt{sentiment\_by}
    \item \texttt{word\_count}, which is the number of words in the given sentence;
    \item \texttt{sentiment}, which is the sentiment score of the given sentence.
\end{enumerate}

The \texttt{extract\_sentiment\_terms()} function helps us extract the keywords - both positive and negative - that led to the sentiment score calculation. The sentimentr package also supports pipe operator $>$ which makes it easier to write multiple lines of code with less assignment and also cleaner code.

And finally, the \texttt{highight()} function coupled with \texttt{sentiment\_by()} gives a HTML output with parts of sentences nicely highlighted with green and red color to show its polarity.

In order to account for valence shifters, the program frames an observation windows of 4 words before and 2 words after the polarized word  to look for valence shifters. Words within this window may be tagged as neutral words, negators, amplifiers, or de-amplifiers. Each polarized word is further weighted by the function and number of the valence shifters directly surrounding the positive or negative word. Amplifiers become de-amplifiers if the observation window contains an odd number of negators. Finally, pause words are considered as breaking the sentence into two pieces.

An example of application of the package to a set of simple sentences is shown in Table \ref{tab:sentimentrexample}. Negators appear to be correctly accounted for, as well as the other valence shifters with two exceptions (third and fourth sentence).

\begin{table}[]
    \centering
    \begin{tabular}{lc}
    \toprule
      Sentence   & Score \\\midrule
      This device is perfect & 0.375\\
       This device is not perfect  & -0.335\\ 
       This device is not perfect and useless & 0\\
       This device is useless and not perfect & 0\\
       This device is not perfect, but useful & 0.142\\
       This device is not perfect, but useless & -0.921\\
       This device is not perfect, and useless & -0.567\\
       This device is useless, and not perfect & -0.567\\
       I like it & 0.289\\
       I do not like it & -0.224\\
       I don't like it & -0.25\\
       I really like it & 0.45\\
       I dislike it & -0.577\\
       I really dislike it & -0.9\\
       \bottomrule
    \end{tabular}
    \caption{Results provided by the \textit{SentimentR} package for sample sentences}
    \label{tab:sentimentrexample}
\end{table}

\section{SentimentAnalysis}
The package is available at \url{https://cran.r-project.org/web/packages/SentimentAnalysis/index.html}. 

It allows to use several specialized existing dictionaries as well as to create new ones. The default dictionaries are (see \cite{kearney2014textual} for survey of specific dictionaries for finance):
\begin{itemize}
    \item The psychological Harvard-IV dictionary (GI), which is a general-
purpose dictionary developed by the Harvard University;
\item Henry's finance-specific dictionary;
\item The Loughran-McDonald finance-specific dictionary;
\item The QDAP dictionary.
\end{itemize}

The Harvard-IV dictionary contains a list of 1316 positive and 1746 negative words according to the psychological Harvard-IV dictionary\footnote{see the page \url{http://www.wjh.harvard.edu/~inquirer/} for more information} as used in the General Inquirer software. Score values are binary.

Henry's finance-specific dictionary contains a list of positive and negative words and can be retrieved through the \textit{loadDictionaryHE()} function. It is quite small, since it contains just 53 positive words and 44 negative words. It was published in 2008 \cite{henry2008investors}.

The Loughran-McDonald finance-specific dictionary contains a list of positive, negative and uncertainty words, and litigious words, and modal words as well. It can be retrieved through the \textit{loadDictionaryLM()} function or on the page \url{https://sraf.nd.edu/textual-analysis/resources/}\footnote{see the page \url{https://www3.nd.edu/~mcdonald/Word_Lists_files/Documentation/Documentation_LoughranMcDonald_MasterDictionary.pdf} for more information on the Loughran-McDonald dictionary}. It contains 145 positive words and 885 negative words. It was published in 2011 \cite{loughran2011liability} \cite{loughranjbf}. 

Finally, the QDAP dictionary contains the list of positive and negative words of the dictionary contained in the qdap package. It contains 1280 positive words and 2952 negative words. It can be retrieved through the \textit{loadDictionaryQDAP()} function.

The sentiment is computed through the \textit{analyzeSentiment()} function, which returns the sentiment scores for each dictionary, separately for both positive and negative categories and on the overall. However, it exhibits the same problems of other packages as to the treatment of negators: the two sentences "This device is good" and "This device is not good" receive the same (positive) sentiment score.

\section{Conclusion}
The four packages under examination differ greatly for the choice of dictionaries, with \textit{SentimentAnalysis} being probably the most specific but also most restricted. However, all offer the choice of creating custom dictionaries.

A critical issue in sentiment computation appears to be the treatment of negators. Actually, all the packages fail to properly account for negators, with the single exception of \textit{SentimentR}, which should then be considered as the package of choice.








\end{document}